%% file: acl_latex.tex
\pdfoutput=1

\documentclass[11pt, dvipsnames]{article}

\usepackage[final]{acl}

\usepackage{times}
\usepackage{latexsym}

\usepackage{microtype}
\usepackage{amsmath}
\usepackage{graphicx}
\usepackage{booktabs}
\usepackage{multirow}
\usepackage{algorithm}
\usepackage{algpseudocode}
\usepackage{tablefootnote}
\usepackage{caption}
\usepackage{subcaption}

\usepackage[T1]{fontenc}

\usepackage[utf8]{inputenc}

\usepackage{microtype}

\usepackage{inconsolata}

\usepackage{graphicx}
\usepackage{authblk}

\title{Fine-Tuning Language Models on Multiple Datasets for Citation Intention Classification}

\author[1]{\bf Zeren Shui}
\author[2]{\bf Petros Karypis}
\author[1]{\bf Daniel S. Karls}
\author[3]{\bf Mingjian Wen}
\author[1]{\\ \bf Saurav Manchanda}
\author[1]{\bf Ellad B. Tadmor}
\author[1]{\bf George Karypis}
\affil[1]{University of Minnesota, MN, USA}
\affil[2]{University of California San Diego, CA, USA}
\affil[3]{University of Houston, TX, USA}

\affil[ ]{\texttt{\{shuix007, karl0100, manch043, tadmor, karypis\}@umn.edu}}
\affil[ ]{\texttt{pkarypis@ucsd.edu}, \texttt{mjwen@uh.edu}}

\begin{document}
\maketitle

\begin{abstract}
Citation intention Classification (CIC) tools classify citations by their intention (e.g., background, motivation) and assist readers in evaluating the contribution of scientific literature. 
Prior research has shown that pretrained language models (PLMs) such as SciBERT can achieve state-of-the-art performance on CIC benchmarks. 
PLMs are trained via self-supervision tasks on a large corpus of general text and can quickly adapt to CIC tasks via moderate fine-tuning on the corresponding dataset. 
Despite their advantages, PLMs can easily overfit small datasets during fine-tuning. 
In this paper, we propose a multi-task learning (MTL) framework that jointly fine-tunes PLMs on a dataset of primary interest together with multiple auxiliary CIC datasets to take advantage of additional supervision signals.
We develop a data-driven task relation learning (TRL) method that controls the contribution of auxiliary datasets to avoid negative transfer and expensive hyper-parameter tuning.
We conduct experiments on three CIC datasets and show that fine-tuning with additional datasets can improve the PLMs' generalization performance on the primary dataset.
PLMs fine-tuned with our proposed framework outperform the current state-of-the-art models by 7\% to 11\% on small datasets while performing competitively with the best-performing model on the largest benchmark dataset.

\end{abstract}

\section{Introduction}

\input{Sections/Introduction}
\section{Related Work}

\input{Sections/Related_Work}

\section{Methodology}

\input{Sections/Multitask_Finetuning}

\section{Experiments}

\input{Sections/Experiments}

\section{Conclusion}

\input{Sections/Conclusion}

\section{Acknowledgement}

\input{Sections/Ackownledgement}

\section{Limitations}

\input{Sections/Limitations}

\section{Ethical Statement}

\input{Sections/Ethical}

\bibliography{custom}

\newpage
\appendix
\input{Sections/Appendix}

\end{document}

%% file: Sections/Introduction.tex
Citation count is a crucial bibliometric for assessing the impact of scientific papers~\cite{manchanda-karypis-2021-evaluating}.
Highly cited papers are often regarded as seminal works in their respective fields.
Scientists cite other papers for various reasons, each contributing differently to the impact of the cited papers.
For example, they may cite a paper because its the bedrock of their work or because it provides background knowledge.
Recently, it is also found that citations can be purchased and manipulated~\cite{ibrahim2024google}.
This highlights the need for tools that can identify the intention behind citations to create more nuanced bibliometrics.






Citation intention classification (CIC) tools classify citations based on their underlying intentions.
Prior research formulates CIC as a text classification problem and solves it using machine learning methods~\cite{jurgens2018measuring, cohan2019structural, berrebbi2022graphcite}. 
They extract the citation context from the citing papers (i.e., a span of text around the citation), and use it as the input to classifiers. 
Among these methods, pretrained language models (PLMs)~\cite{devlin2019bert, liu2019roberta} achieve the current state-of-the-art performance on CIC benchmarks~\cite{beltagy2019scibert}. 
Researchers apply PLMs to the CIC problem by fine-tuning them on citations with intention labels.
However, obtaining labeled citations is challenging, as labeling citations in a scientific domain requires experts with in-depth domain knowledge.

Over the years, different CIC datasets have been curated that assign the citations to different intention categories~\cite{hernandez2016citationsurvey}. 
They share an input space which is a set of citation contexts extracted from scientific papers.
The set of citation labels of these datasets may contain semantically identical or similar intention categories (e.g., \emph{"Background"} and \emph{"Unused"}).
Accordingly, fine-tuning a PLM on one such dataset may benefit its generalization performance on others.
In this paper, we aim to improve PLMs' generalization performance on CIC datasets by leveraging supervision signals from additional CIC datasets.

We propose a multi-task learning (MTL) framework that jointly fine-tunes PLMs on auxiliary CIC datasets to improve the PLMs' generalizability on a primary CIC dataset of interest. 
To prevent negative transfer~\cite{wang2019negtrans1} (wherein sharing information with unrelated tasks harms the performance of the primary task) and to reduce the burden of hyper-parameter search, we propose a task relation learning (TRL) method to control the contribution of the auxiliary datasets to the training.
The TRL method measures the relevance of an auxiliary dataset to the primary dataset by evaluating the \emph{information gain}~\cite{shannon1948mathematical} of a model trained on the auxiliary dataset on the primary dataset. 
We also find that the position of a citation within its context provides useful information for CIC tasks and a position-aware readout function, i.e., a function that aggregates the PLM output token embeddings to a fixed length citation embedding, can improve PLMs' performance.

Our contributions are summarized as follows: 
\begin{itemize}
    \item We introduce a MTL framework that fine-tunes PLMs jointly on multiple CIC datasets to improve their generalizability on the primary dataset of interest. 
    \item We propose a data-driven TRL method that controls the contribution of auxiliary datasets in the MTL framework. It effectively and efficiently avoids negative transfer.
    \item We find that the position of the citation within the context is an informative signal for predicting citation intentions. We propose a position-aware readout function that outperforms the commonly used CLS and MEAN readout functions. 
    \item We curate a new benchmark dataset called KIM that is specialized for the development of CIC applications in materials science. 
\end{itemize}
We carefully design and conduct experiments on three benchmark datasets which show that jointly fine-tuning PLMs on multiple datasets with the proposed MTL framework improves the PLMs' performance on the primary dataset. 
PLMs fine-tuned with our framework outperform the current state-of-the-art models by 7\% to 11\% on small datasets while align with the best-performing model on a large dataset. 
We release the code and datasets used in our experiments at \url{https://github.com/shuix007/Deep-Citation.git}.

%% file: Sections/Related_Work.tex
\subsection{Citation Intention Classification}

CIC is a classification task that assigns citations into discrete intention categories such as \textit{background} and \textit{motivation}. 
A citation consists of several components, including a citation context (i.e., a span of text that contains the citation), topology information about the citing paper and the cited paper (e.g., their neighbors in citation graphs), meta information such as the title of the section that contains the citation, and etc.

Citation context is arguably the primary signal for CIC methods. The majority of prior research formulates CIC as a text classification problem and focuses on featurization of citation contexts. 
Early works~\cite{abu-jbara-etal-2013-purpose, jurgens2018measuring} represent citation contexts by hand-engineered features and pre-trained word embeddings (e.g., GloVe~\cite{pennington-etal-2014-glove}, ELMo~\cite{DBLP:journals/corr/abs-1802-05365}). 
These methods apply traditional classification models such as support vector machines (SVM)~\cite{scholkopf2002SVM} to predict citation intentions. 
Deep learning-based methods~\cite{cohan2019structural} use word embeddings together with a bi-directional long short-term memory network (BiLSTM)~\cite{hochreiter1997lstm} to learn context representations end-to-end for CIC tasks. 
In recent years, Transformer~\cite{vaswani2017attention}-based PLMs~\cite{devlin2019bert,liu2019roberta} revolutionized a wide range of NLP tasks, including CIC. 
PLM-based methods that fine-tune different PLMs such as SciBERT~\cite{beltagy2019scibert} and XLNet~\cite{yang2019xlnet} achieve the state-of-the-art performance on various CIC benchmarks~\cite{mercier2020impactcite,lahiri2023citeprompt}.

Some methods explore leveraging the other sources of information to improve CIC performance.
\citet{cohan2019structural} demonstrate that training CIC models with two auxiliary tasks, citation worthiness prediction and the section title prediction, effectively improves the generalization performance of CIC models. 
\citet{berrebbi2022graphcite} leverage the topology information from citation graphs as additional signals for predicting citation intentions and achieve better performance than methods that only predict by citation context.

\begin{figure*}[t]
\centering
\includegraphics[width=1.0\textwidth]{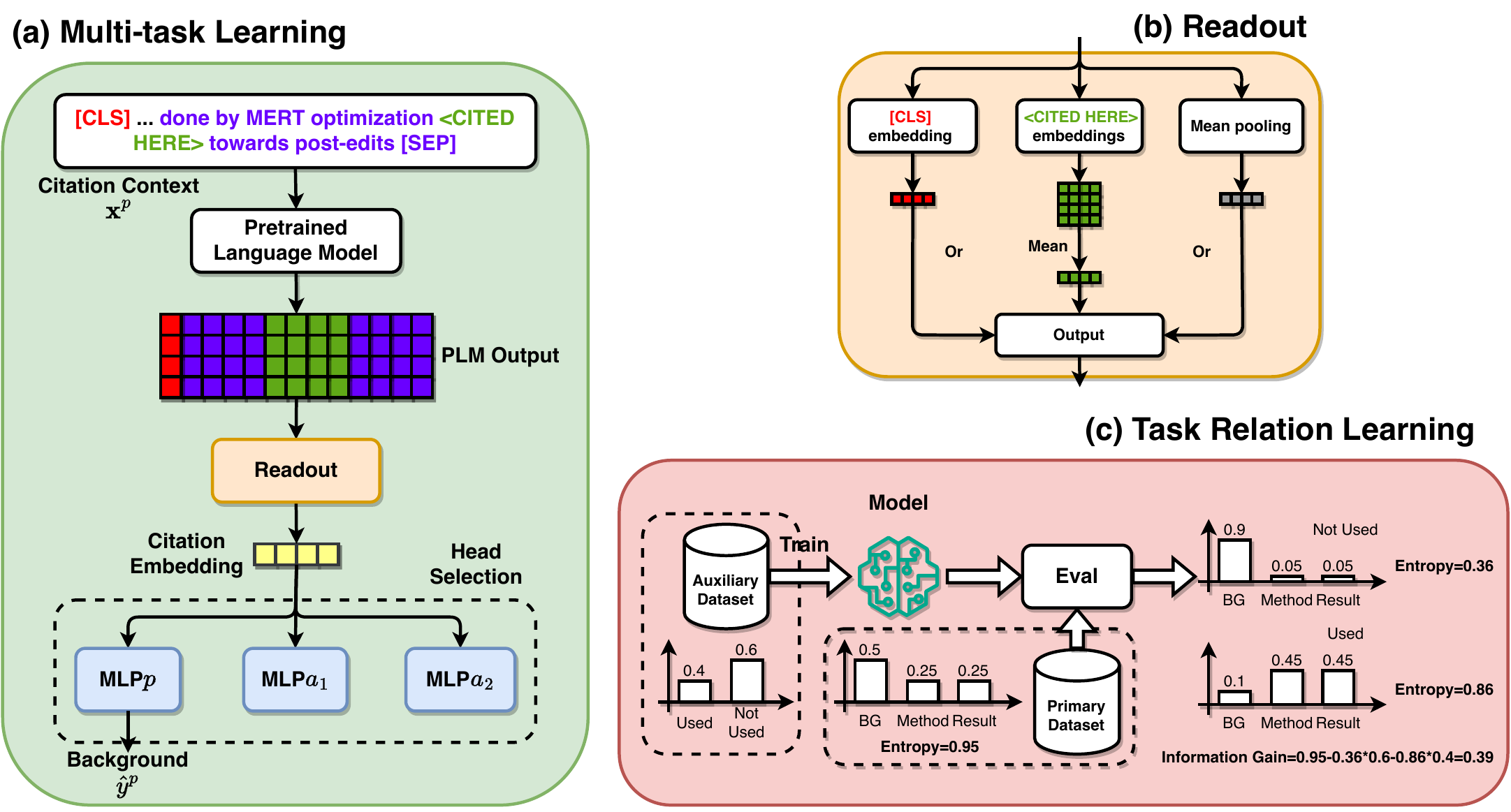}
\caption{Overview of the architecture of our multi-task learning framework. (a) An overview of the MTL training process. The same language model parameters are shared across all datasets.; (b) The three readout operations (CLS, MEAN, CITED HERE) over the language model embeddings to generate representations for citation contexts; (c) The task relation learning (TRL) method. We train a classification model on an auxiliary dataset then evaluate its information gain on the primary dataset.}
\label{fig:DeepCite}
\end{figure*}

\subsection{Multi-task Learning}

Multi-task learning (MTL)~\cite{ruder2017mtl} is a paradigm that jointly trains a model with shared parameters on multiple related tasks such that the knowledge acquired from them can benefit the learning of each other~\cite{yu2020bdd100k,liu-etal-2019-multi,tao2020end}.
MTL formulations usually fall into two categories. 
The first category, which includes the present work, has one or more primary tasks (a.k.a.\ target tasks) and a set of auxiliary tasks that serve as regularizers to improve the model's generalizability on the primary tasks~\cite{ning2009multisar,liebel2018auxiliary,cohan2019structural}.
In the second category, tasks are equally important and the goal is to reach the Pareto frontier~\cite{lin2019pareto} over these tasks~\cite{cao2022relational,kendall2018multi}. 

Task weight assignment is an essential topic in MTL research. 
Earlier methods either assign task weights apriori by experts to reflect domain preference~\cite{kokkinos2017ubernet,eigen2015predicting} or tune the weights as hyper-parameters using validation sets~\cite{cohan2019structural}. \citet{kendall2018multi} propose to weigh different tasks by a measurement of their uncertainties and show improved performance on all tasks compared to training separate models for each task. 
Our task relation learning method is a data-driven method that finds the relation between each auxiliary dataset (task) and the target dataset (task) to improve the model's performance on the target dataset.

\subsection{Pretrained Language Model}

Pretrained language models (PLM), a.k.a. large language models (LLM), are models pretrained on unlabeled text corpus with self-supervised language modeling tasks such as causal language modeling~\cite{radford2018gpt1} and masked language modeling~\cite{taylor1953cloze}. PLMs achieve state-of-the-art performance on a wide range of downstream NLP tasks via gradient-based fine-tuning on the corresponding datasets~\cite{devlin2019bert, liu2019roberta, beltagy2019scibert, raffel2020t5, radford2019gpt2}.

Recently, ~\citeauthor{brown2020gpt3} demonstrate generative PLMs can be specialized for different tasks with in-context learning (ICL), a fine-tuning-free learning paradigm that unifies NLP tasks to the language generation task. Although the increasing scale of PLMs has been improving the performance of ICL on NLP tasks~\cite{wei2023larger}, we find that ICL with GPT4~\cite{bubeck2023gpt4} still underperforms our proposed method.

%% file: Sections/Multitask_Finetuning.tex


\subsection{Problem Setting} \label{sec:CIC_def}

We formulate the CIC task as a text classification problem. 
A CIC dataset $\mathcal{D}^{t} = \{\mathbf{x}_i^t, y_i^t\}_{i=1}^{|\mathcal{D}^t|}$ is a set of instances where $x_i^t \in \mathcal{X}^t$ is a citation context and $y_i^t \in \mathcal{Y}^t$ is a discrete variable that indicates the intention of the citation. 
A citation context is a span of text around the citation.
We posit the existence of multiple CIC datasets and use the superscript $t$ to distinguish different datasets.
In the MTL setting, there is a primary dataset of interest $\mathcal{D}^p$ and a set of auxiliary datasets $\{\mathcal{D}^a \}_{a \in A}$. 
We assume the datasets share an input space $\mathcal{X}^t = \mathcal{X}, \forall t \in \{p\} \cup A$ (the text of the citation context from scientific literature).
We do not posit any assumptions on the label space $\mathcal{Y}^t$, different CIC datasets can have different label spaces.
Our goal is to train a CIC model $\mathcal{M}$ such that
\[
    \mathcal{M}\left(\mathbf{x}_i^p \right) = y_i^p, \forall \left(\mathbf{x}_i^p, y_i^p \right) \in \mathcal{D}^p.
\]

\subsection{Multi-task Learning} \label{sec:mtl}


Although the intention (label) spaces for different CIC datasets are different, most of the CIC datasets share an input space, citation context from scientific literature, for predicting the intention of the citations~\cite{hernandez2016citationsurvey}. 
Moreover, the label spaces of the datasets may contain semantically similar or even shared intention categories (e.g., "Background" and "Not Used").
It is intuitive to assume leveraging the supervision signal of one dataset to fine-tune a PLM can improve its performance on another dataset.
To that end, we propose a MTL framework that jointly fine-tunes a PLM on additional CIC datasets as auxiliary tasks to improve its generalizability on a primary dataset.

The MTL framework shares a PLM across datasets while using a separate prediction head (MLP) for each of them. 
During training, each dataset-specific MLP optimizes its parameters using gradients computed from its own prediction while losses of all prediction heads back-propagate to the PLM to update its parameters. 
Inference over the primary dataset is performed using its corresponding MLP. The MTL framework is shown in Figure~\ref{fig:DeepCite}.



Letting $\Theta_{\text{LM}}$ be the parameters of the PLM and $\Theta_{\text{MLP}_a}$ be the parameters of the MLP for dataset $a$, the objective function for the MTL framework is 
\begin{equation} \label{eq:mhmtl}
\begin{aligned}
    \mathcal{L} = & \frac{1}{|\mathcal{D}^p|} \sum_{(x_i^p, y_i^p) \in \mathcal{D}^p} l^p \left( f^p\left( x_i^p; \Theta_{\text{LM}}, \Theta_{\text{MLP}_p} \right), y_i^p \right) \\
    & + \sum_{a \in A} \frac{\lambda_a}{|\mathcal{D}^a|} \sum_{(x_i^a, y_i^a) \in \mathcal{D}^a} l^a \left( f^a\left( x_i^a; \Theta_{\text{LM}}, \Theta_{\text{MLP}_a} \right), y_i^a \right)
\end{aligned}
\end{equation}
where $p$ is the primary dataset and $A$ is the set of auxiliary datasets. $f^a(\cdot)$ and $l^a(\cdot)$ are the prediction function and the cross-entropy loss associated with dataset $a$, respectively. The coefficients $\lambda_a \in [0, 1]$, whose importance we discuss in the next section, control the contribution of dataset $a$ towards the primary dataset $p$.

\subsection{Task Relation Learning} \label{sec:trl}

In MTL, sharing information with relevant tasks may benefit the primary task but learning unrelated tasks may harm the performance of the primary task (negative transfer)~\cite{wang2019negtrans1}. Identifying related tasks is thus critical in designing MTL frameworks. As shown in Equation~\ref{eq:mhmtl}, we use a coefficient $\lambda_a$ to control how much an auxiliary dataset $a$ contributes to the primary dataset. A common way to determine the value of $\lambda_a$ is to treat it as a hyper-parameter and tune its value for best primary accuracy on a validation set via grid search. However, the search space grows exponentially as the number of auxiliary datasets increases, making a grid search too inefficient.

Since all datasets are for CIC tasks and share input space, a model trained on one dataset may generate meaningful predictions in the label space of another dataset. For example, a citation that is labeled as \emph{"Background"} in one dataset may be classified as \emph{"Unused"} by a model trained on another dataset. 
Based on this insight, we propose a task relation learning (TRL) method to determine the value of $\lambda_a$. 
Our method trains a model on the auxiliary dataset $a$ and evaluates it on the primary dataset $p$. 
If the model performs well on the primary dataset, we assume that jointly training on the auxiliary dataset could benefit the model's performance on the primary dataset.
Considering the different label spaces of the datasets, it is intractable to employ traditional classification metrics such as accuracy and F1 scores for such evaluations. 
We propose to use information gain as the metric for the evaluation.




Information gain is defined on the basis of entropy~\cite{shannon1948mathematical}, a measure of uncertainty in information theory. Let $\mathcal{Y}^p$ and $\mathcal{Y}^a$ be the label spaces of the primary dataset and an auxilary dataset, respectively. The entropy of $\mathcal{Y}^p$ is
\begin{equation}
\text{Entr}(\mathcal{Y}^p) = - \sum_{i \in \mathcal{Y}^p} P(i) \log_{|\mathcal{Y}^p|} (P(i)),
\end{equation}
which measures the uncertainty in set $\mathcal{Y}^p$ in the range of $[0, 1]$. $P(i)$ is the probability that intention $i$ emerges in the dataset. When there is only one label $i \in \mathcal{Y}^p$ in the dataset, it is certain that any sample will have label $i$ and the entropy is zero. When the labels are evenly distributed, a sample from the dataset is completely uncertain and the value of the entropy is one. We compute entropy on the label distribution of the primary dataset.

After training a model on the auxiliary dataset, we apply the model to the primary dataset and obtain a predicted label from the auxiliary label space for each instance. We group the instances by their predicted labels and obtain a conditional primary label distribution for each auxiliary label $j \in \mathcal{Y}^a$. 
We compute entropy on each conditional label distribution by
\begin{equation}
\text{Entr}(\mathcal{Y}^p|j) = - \sum_{i \in \mathcal{Y}^a} P(i|j) \log_{|\mathcal{Y}^p|} (P(i|j)),
\end{equation}
where $P(i|j)$ denotes the probability that a primary instance with label $i$ is predicted into the auxiliary intention $j$. Note that, a small value of $\text{Entr}(\mathcal{Y}^p|j)$ indicates the predicted auxiliary label $j$ is likely to be a sub-class of one label in the primary label space $\mathcal{Y}^p$ as most of the instances with predicted label $j$ belong to one label in the primary label space $\mathcal{Y}^p$. 
We calculate information gain by
\begin{equation}
    \text{IG}(\mathcal{Y}^p|\mathcal{Y}^a) = \text{Entr}(\mathcal{Y}^p) - \sum_{j \in \mathcal{Y}^a} P(j) (\text{Entr}(\mathcal{Y}^p|j)),
\end{equation}
where $P(j)$ is the probability that a primary instance is predicted as label $j$. 
Information gain quantifies the uncertainty reduction when we group the primary instances by their predicted label in the auxiliary label space. A large value of information gain indicates that each label in the auxiliary space is a sub-class of a primary label which means the auxiliary label space is highly correlated with the primary label space.

We compute the value of $\lambda_a$ as the relative reduction of entropy
\begin{equation}
    \lambda_a = \frac{\text{IG}(\mathcal{Y}^p|\mathcal{Y}^a)}{\text{Entr}(\mathcal{Y}^p)},
\end{equation}
which falls into the range of $[0, 1]$. The closer $\lambda_a$ is to one, the more similar dataset $a$ is to the primary dataset. We show an example of the TRL method in Figure~\ref{fig:DeepCite}.

\subsection{Readout Function} \label{sec:backbone}



A common practice of fine-tuning PLMs for text classification is to use them as text encoders that convert citation contexts to vectors in a latent space and feed the latent vectors through prediction heads (MLPs) to obtain classification probabilities~\cite{sun2019finetunebert}. 
PLMs output a contextualized embedding for tokens at each position of the citation contexts. 
For each context, we use a readout function to aggregate its contextualized token embeddings to a sentence embedding to feed into the downstream MLPs. 
We explore two standard readout functions, CLS that uses the output embedding of the <CLS> token as the context embedding, and MEAN that averages the contextualized embeddings of all tokens to be the context embedding~\cite{beltagy2019scibert, devlin2019bert, reimers2019sbert}. 
We propose a third approach called CITED HERE that is motivated by the fact that the position of the citation in the citation context can be informative for predicting the intention of the citation~\cite{jurgens2018measuring}.
We insert a special mark, <CITED HERE>, into the position of the citation in the context and apply mean pooling on the embeddings of the corresponding tokens as the representation of the citation to feed into MLPs (see an example in Figure~\ref{fig:DeepCite}). 
Our experiments in Section~\ref{sec:readout} show that the CITED HERE readout function performs better than the standard position-agnostic readout functions.




%% file: Sections/Experiments.tex


\begin{table} 
  \centering
  \begin{tabular}
  {lrrr}
    \toprule
    Dataset & \# instances & \# papers \tablefootnote{Number of unique citing papers.} & \# labels \\
    \midrule
    ACL & 1904 & 186 & 6   \\
    KIM  & 804 & 614 & 3  \\
    SciCite  & 11020 & 6627 & 3  \\
  \bottomrule
\end{tabular}
\caption{Statistics of the datasets}
\label{tab:data_stat}
\end{table}

\subsection{Datasets}

We conduct experiments on three datasets: \textbf{ACL}~\cite{jurgens2018measuring}, \textbf{SciCite}~\cite{cohan2019structural}, and a newly curated in-house dataset called \textbf{KIM}. 
The ACL dataset was collected from the ACL Anthology Reference Corpus and consists of natural language processing papers.
The SciCite dataset is the largest one we consider and contains citations from general computer science and medical domains. The papers are collected from the Semantic Scholar Open Research Corpus~\cite{lo-etal-2020-s2orc}. 
Detailed statistics and descriptions of the datasets are shown in Table~\ref{tab:data_stat} and Table~\ref{tab:intention_def}. 

\paragraph{KIM Dataset.} Despite the fact that ACL and SciCite are widely used as CIC benchmarks, their focus on the fields of computer science and medicine renders them insufficient for building CIC models applicable to other scientific domains.
In this paper, we curate a new CIC dataset in the field of materials science called KIM.
The KIM dataset was constructed by collecting the primary citations for interatomic models archived in the OpenKIM repository\footnote{openkim.org}~\cite{tadmor2011kim}, retrieving as many papers as possible from the literature
that cite any one of them, and extracting the associated citation context(s). 
This forms a set of 804 citations to be labeled.
The guidelines describing each of the annotation labels assigned for the KIM dataset are provided in Table~\ref{tab:intention_def}. The KIM dataset was annotated by three different domain experts but the labeling was not performed independently. While the initial labeling was carried out separately on disjoint subsets of the full dataset, all annotators ultimately reviewed each and every label together and came to an agreement for it.

\begin{table}
  \resizebox{0.5\textwidth}{!}{
  \centering
  \begin{tabular}
  {llccc}
    \toprule
    PLM & Method & ACL & KIM & SciCite \\
    \midrule
     & Scaffolds & 67.90 & - & 84.00  \\
    \midrule
    \multirow{2}{*}{GPT4} & ICL 0-shot & 38.55 & 33.86 & 72.86 \\
    & ICL 5-shot & 50.18 & 60.55 & 74.55 \\
    \midrule
    XLNet & ImpactCite & 64.62 & 61.01 & 84.98 \\
    \midrule
    \multirow{3}{*}{BERT} & Default &  57.44 & 57.30 & 83.46 \\
                        & Ours (Search) & 65.98 & \underline{64.18} & 84.08 \\
                        & Ours (TRL) & 66.32 & 62.00 & 83.48 \\
    \midrule
    \multirow{4}{*}{SciBERT}  & Default & 67.25 & 60.27 & 85.22 \\
                        & CitePrompt & 66.58 & 62.22 & 85.02 \\
                        & Ours (Search) & \underline{73.74} & 63.11 & \underline{85.25} \\
                        & Ours (TRL) & \textbf{75.57} & \textbf{64.56} & \textbf{85.35} \\
  \bottomrule
\end{tabular}
}
\caption{Performance (Macro-F1) of the MTL fine-tuning approach compared to the baseline methods. Search indicates grid search while TRL indicates our proposed task relation learning method. Results are averaged over five runs, the best performing method for each primary dataset is in \textbf{bold}, and the second best results are \underline{underlined}.}
\label{tab:overall_results}
\end{table}

\begin{figure*}[t]
\centering
\includegraphics[width=\textwidth]{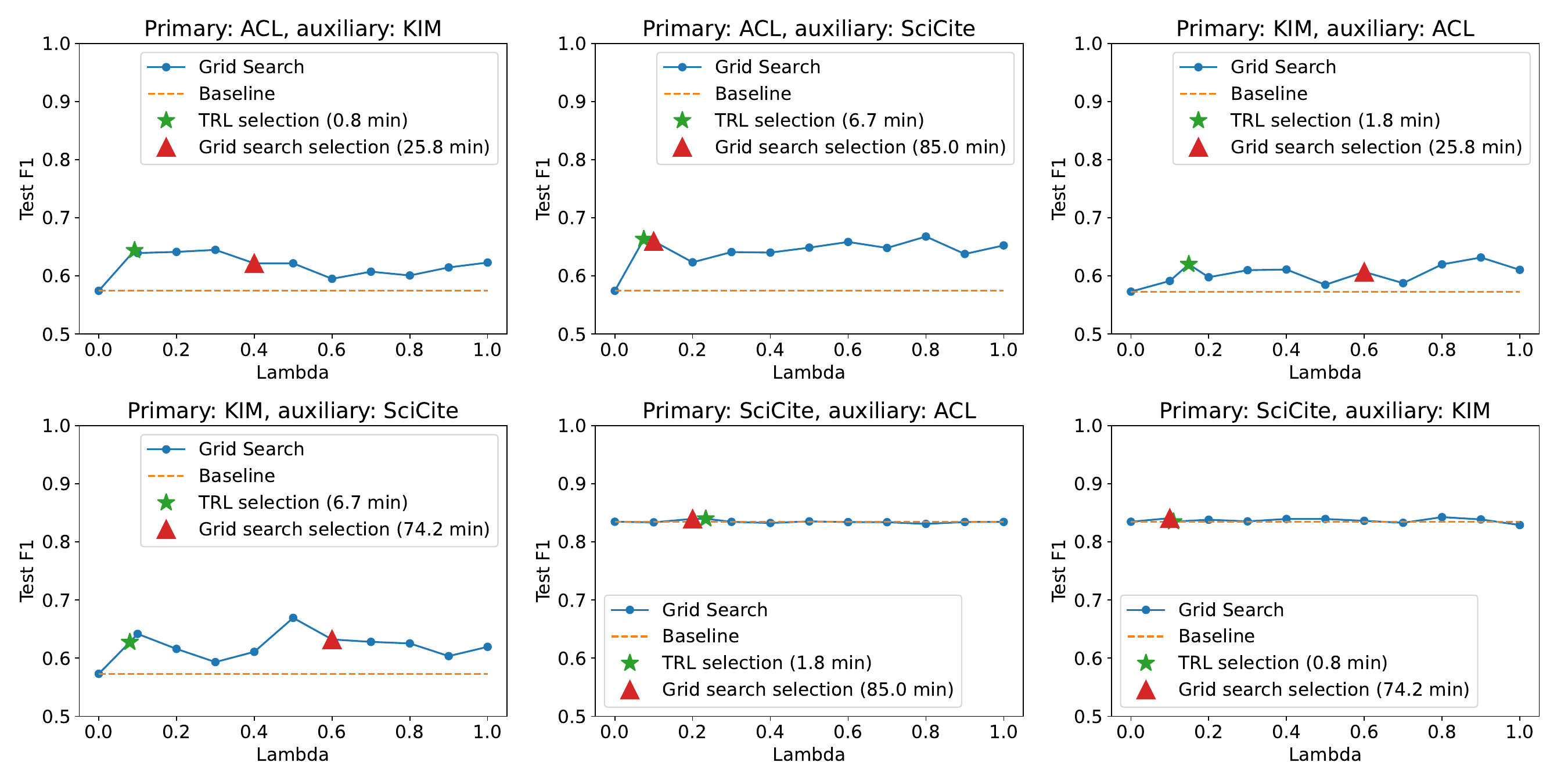}
\caption{BERT performance of all binary combinations of primary and auxiliary datasets with different value of $\lambda$s. The yellow line denotes the baseline performance of fine-tuning on only the primary dataset. The blue line denotes the performance of fine-tuning the primary and the auxiliary dataset with different $\lambda$s. The star and the triangle indicate the $\lambda$ found by our TRL method and the grid search method, respectively. Time in the brackets indicates the GPU time needed for the method.}
\label{fig:trl_lambdas}
\end{figure*}

\subsection{Baselines}

We compare our methods with \textbf{Scaffolds}~\cite{cohan2019structural}, \textbf{ImpactCite}~\cite{mercier2020impactcite}, \textbf{CitePrompt}~\cite{lahiri2023citeprompt}, \textbf{BERT}~\cite{devlin2019bert}, \textbf{SciBERT}~\cite{beltagy2019scibert}, and \textbf{GPT4}~\cite{bubeck2023gpt4}. 
Scaffolds is the state-of-the-art RNN-based CIC method that does not rely on PLMs. 
We report its results from the original paper for comparison as we use the same train-test split of the ACL and the SciCite dataset. 
ImpactCite and CitePrompt are two PLM-based CIC models. ImpactCite fine-tunes XLNet~\cite{yang2019xlnet} while CitePrompt apply prompt-tuning methods to SciBERT for CIC tasks.
For these two baseliens, we use the codebases and the training configurations provided by the authors.
BERT and SciBERT are PLMs pretrained on general domain text corpus and scientific literature, respectively. For BERT and SciBERT as baselines, we follow the setting from their original papers and use the output embedding of the CLS token as the context representation.

We evaluate zero-shot and few-shot ICL performance on one of the most capable generative PLMs, GPT4. We prompt GPT4 with a detailed text description of the CIC task, definition of the intentions, and/or a few examples. For few-shot experiments, we follow the common practice in ICL~\cite{brown2020gpt3} and randomly select five examples for each intention class as examples. Details about the prompts are shown in Section~\ref{sec:icl}.

\subsection{Experimental Settings}

We evaluate our proposed multi-task learning (MTL) framework, the task relation learning (TRL) method, and the position-aware CITED HERE readout function on two backbone PLMs, BERT and SciBERT. 
We employ two methods to compute the value of the aforementioned $\lambda$ coefficients: grid search and the TRL method. For grid search, we explore values in the range $0.1$ to $1.0$ in increments of $0.1$ using the validation set of the primary dataset. 
For the TRL method, we fine-tune one PLM on an auxiliary dataset and evaluate its information gain on the training set of the primary dataset to compute a $\lambda$ that is associated with the PLM-primary-auxiliary triplet. 
This $\lambda$ is used for jointly fine-tuning the PLM that computes it on the primary-auxiliary datasets. 
For fine-tuning a PLM on a primary dataset with more than one auxiliary dataset, we use the $\lambda$s associated with the PLM-primary-auxiliary triplets, respectively.
In all MTL experiments involving BERT and SciBERT, CITED HERE is the default readout function unless stated otherwise. A detailed experimental setting is shown in Section~\ref{sec:exp_setting}.


\subsection{Results}

The main results of our experiments can be found in Table~\ref{tab:overall_results}. 
Our method achieves the state-of-the-art performance on the three benchmark datasets.
On the two small benchmark datasets, ACL and KIM, our proposed fine-tuning framework significantly improves the backbone PLMs' performance compared to the default fine-tuning process. In particular, our method outperforms the current state-of-the-art by 7\% on the KIM dataset and by 11\% on the ACL dataset. 
On the largest benchmark dataset of the three, SciCite, our framework perform competitively with the best-performing baselines. 
Our method outperforms the zero-shot and few-shot ICL methods on GPT4 by a significant margin. This demonstrates that, despite the increasing reasoning capability of LLMs, in the CIC application with a few thousands training instances, our methodology is still necessary.

\begin{figure*}[t]
\centering
\includegraphics[width=\textwidth]{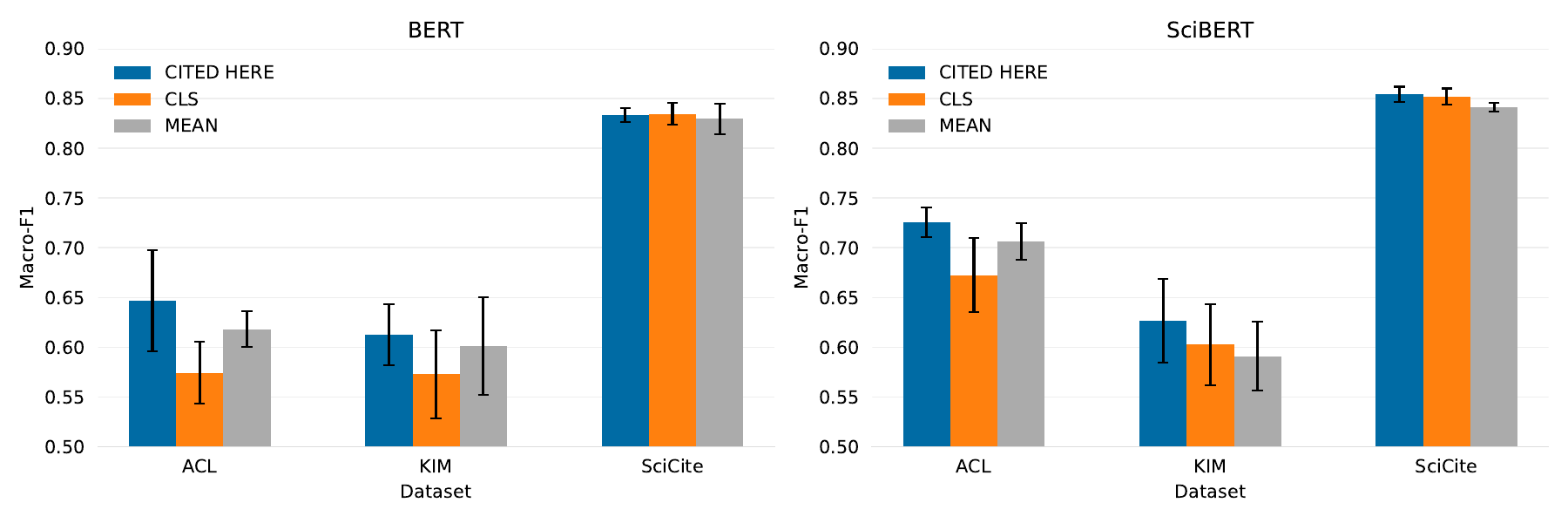}
\caption{Performance (Macro-F1 with standard deviation) of the three readout functions: CLS, MEAN, and CITED HERE on BERT and SciBERT.}
\label{fig:readout}
\end{figure*}

\subsection{Task Relation Learning}

We investigate the effectiveness of our proposed TRL method for identifying the value of the $\lambda$ coefficients. 
In Figure~\ref{fig:trl_lambdas}, we show BERT's performance of all binary combinations of primary and auxiliary datasets with different value of $\lambda$s. 
For each pair of primary-auxiliary datasets, we show the $\lambda$ values identified by our TRL method and the grid search method. 
We observe that the choice of $\lambda$ is critical to the joint fine-tuning performance. A poorly chosen $\lambda$ could amortize the benefit of auxiliary datasets and even degrade the performance.

When the auxiliary dataset can lead to positive transfer, our TRL method can effectively identify $\lambda$s that improve the performance of the primary datasets and it performs on par or better than those selected by the grid search method (e.g., Primary: KIM, Auxiliary: SciCite). 
On the other hand, when the primary dataset causes negative transfer, the TRL method chooses a small value of $\lambda$ that avoids performance degradation on the primary dataset (e.g., Primary: ACL, Auxiliary: KIM). We reach similar conclusions on SciBERT and show the analysis in Section~\ref{sec:add_exp_res}.

Note that, the TRL method is also significantly faster compared to the grid search method, exhibiting a factor of 10 to 100 in run time improvement when only one auxiliary dataset is used.
This advantage is more significant when dealing with more auxiliary datasets.




\subsection{Readout Function} \label{sec:readout}

We compare the performance of the three readout functions, CLS, MEAN and CITED HERE and show the results in Figure~\ref{fig:readout}. 
On the two small datasets ACL and KIM, the CITED HERE readout function significantly outperforms the other two readout functions. 
On the SciCite dataset, CITED HERE matches the performance of CLS and MEAN on BERT and slightly outperforms them on SciBERT.

This demonstrates that the position of the citation in the citation context is informative for the prediction of the its intention. Our proposed position-aware readout function, CITED HERE, has an edge over position-agnostic readout functions especially on small datasets. 

\subsection{Ablation on Dataset Size}

To investigate the influence of dataset size on the effectiveness of our method, we conduct additional experiments using subsets of various sizes from SciCite.
In the first experiment, we use SciCite subsets (20\% and 50\%) as the primary datasets and the ACL and KIM datasets as auxiliary datasets. 
We then apply our method to fine-tune SciBERT on these datasets.
The results in Table~\ref{tab:scicite_primary} indicate that the performance improvement is more significant on the subsets of SciCite compared to the entire SciCite dataset.
This demonstrates that smaller primary datasets benefit more from our method and the additional supervision signals from the auxiliary datasets. In the second experiment, we use the ACL and the KIM datasets as the primary datasets and SciCite subsets as the auxiliary datasets.
We observe from Table~\ref{tab:scicite_auxiliary} that, our method’s performance increases significantly as the size of the auxiliary dataset grows.





\begin{table}
  \resizebox{0.5\textwidth}{!}{
  \centering
  \begin{tabular}
  {lrrr}
    \toprule
     & 20\% (1.5K) & 50\% (3.9K) & 100\% (7.7K) \\
    \midrule
    Baseline & 83.87 & 84.28 & 85.22  \\
    Ours (TRL) & 85.02 (\small +1.15) & 85.44 (\small +1.16) & 85.35 (\small +0.13) \\
  \bottomrule
\end{tabular}
}
\caption{Performance (Macro-F1) of SciBERT fine-tuned using the default method and our proposed MTL + TRL fine-tuning approach on a 20\% subset, a 50\% subset of SciCite, and the full SciCite dataset.}
\label{tab:scicite_primary}
\end{table}

\begin{table}
  \resizebox{0.5\textwidth}{!}{
  \centering
  \begin{tabular}
  {lrrrr}
    \toprule
     & Baseline & 5\% (0.4K) & 20\% (1.5K) & 100\% (7.7K) \\
    \midrule
    ACL & 67.25 & 74.56 & 75.35 & 75.57  \\
    KIM & 60.27 & 62.35 & 63.47 & 64.56 \\
  \bottomrule
\end{tabular}
}
\caption{Performance (Macro-F1) of the proposed MTL + TRL fine-tuning approach when using ACL and KIM as the primary dataset, respectively. The auxiliary datasets are subsets of three different sizes of SciCite (5\%, 20\%, and 100\%).}
\label{tab:scicite_auxiliary}
\end{table}

%% file: Sections/Conclusion.tex
We propose a multi-task learning (MTL) framework to fine-tune pretrained language models (PLMs) for citation intention classification (CIC) tasks. 
Our framework treats additional CIC datasets as auxiliary tasks to be jointly trained with a primary CIC dataset. 
We develop an efficient, data-driven task relation learning (TRL) method that controls the contribution of auxiliary datasets to avoid negative transfer.
The proposed TRL method effectively identifies a set of coefficients that is critical to the performance of the MTL framework with magnitudes lower computational cost compared to grid search.
We introduce a position-aware readout function and demonstrate that a citation's position within the context is informative for predicting its intention.
Experimental results suggest that jointly fine-tuning PLMs on primary and auxiliary datasets with our proposed MTL framework effectively improves their performance on the primary datasets.


%% file: Sections/Ackownledgement.tex
This work was supported in part by NSF (1447788, 1704074, 1757916, 1834251, 1834332), Army Research Office (W911NF1810344), the startup funds from the Presidential Frontier Faculty Program at the University of Houston, Intel Corp, and Amazon Web Services. Access to research and computing facilities was provided by the Minnesota Supercomputing Institute. OpenKIM acknowledges the support of the Allen Institute for AI through the Semantic Scholar project for providing citation information and full text of articles when available, which are used to train the Deep Citation ML algorithm. We thank the anonymous reviewers for their feedback during the review process.

%% file: Sections/Limitations.tex
In this paper, we experiment with one way of finetuning PLMs, i.e., finetuning the pretrained PLM encoder with randomly initialized task specific multi-layer perceptrons while the multi-task learning (MTL) framework and the task relation learning (TRL) method proposed in this work are supposed to be applicable to any classification models. 
In the future, we will extend our study to different PLM finetuning techniques such as soft-prompt tuning~\cite{lester2021prompttuning} and adaptor-based finetuning~\cite{houlsby2019adapter, hu2021lora} and different CIC models such as GraphCite~\cite{berrebbi2022graphcite}.

%% file: Sections/Ethical.tex
Our work aims to improve the accuracy of citation intention classification (CIC) tools that assist readers in comprehending scientific literature and evaluate the relevance and contribution of scientific publications. The method could be extended to other classification tasks. The datasets we used in this work are generated from scientific literature that are accessible through scientific publishers or pre-print servers. We believe our work should not raise any ethical concerns.

%% file: Sections/Appendix.tex
\begin{table*}[h]
  \resizebox{\textwidth}{!}{
  \begin{tabular}
  {lll}
    \toprule
    Dataset & Intention & Definition \\
    \midrule
    \multirow{6}{*}{ACL-ARC} & Background & The citation provides relevant information for the domain that the present paper discusses.  \\
                        & Motivation & The citation illustrates the need for data, goals, methods, etc that is proposed in the present paper. \\
                        & Uses & The present paper uses data, methods, etc., from the paper associated with the citation. \\
                        & Extends & The present paper extends the data, methods, etc. from the paper associated with the citation. \\
                        & Compare or Contrast & The present paper expresses similarity / differences to the citation. \\
                        & Future & The citation is a potential avenue for future work of the present paper. \\
    \midrule
    \multirow{3}{*}{KIM}  & Used & The present paper uses at least one method that is proposed in the paper associated with the citation. \\
                        & Not Used & The present paper does not use or extend any methods that is proposed in the paper associated with the citation. \\
                        & Extended & The present paper uses an extended / modified version of the method proposed in the paper associated with the citation. \\
    \midrule
    \multirow{4}{*}{SciCite}  & \multirow{2}{*}{Background} & The citation states, mentions, or points to the background information giving more context about a problem, \\ 
    & & concept, approach, topic, or importance of the problem that is discussed in the present paper. \\
                        & Method & The present paper uses a method, tool, approach or dataset that is proposed in the paper associated with the citation. \\
                        & Result & The present paper compares its results/findings with the results/findings of the paper associated with the citation. \\
  \bottomrule
\end{tabular}}
\caption{Definition of intentions.}
\label{tab:intention_def}
\end{table*}

\section{Intention Definitions}

We show the definition of the intentions of the three CII datasets we used in our experiments in Table~\ref{tab:intention_def}. Although the label spaces of the three datasets are different, they contain semantically similar and shared intention categories. For example, "Background" exists in both the ACL-ARC and the SciCite dataset. "Method" in the SciCite dataset and "Used" in the KIM dataset are semantically similar to each other.

\section{KIM Dataset}

\subsection{Choice of the Labels}

As listed in Table 4, each data point in the KIM dataset was assigned to one of three labels: “Used”, “Not Used”, and “Extended”. A label of “Used” indicates that the exact model presented in the cited paper is used in materials simulations in the citing paper without modification; “Extended” means that the model of the cited paper is updated or built upon in some aspect before subsequently used in materials simulations carried out in the citing paper; “Not Used” means that the model is not used in any of the simulations of the citing paper but rather that the citation provides other information such as background and motivation. The reasoning for having only these three labels is domain-specific. A typical material scientist evaluating the influence of an interatomic potential model on a literary work would typically only discern between these labels—any finer granularity is irrelevant.

\subsection{Motivation of the Dataset}

Because CIC models are typically trained on the same widely available datasets pertaining only to several fields, specifically computer science, the introduction of a dataset from a new domain to the study provides a way to better evaluate their transferability. A model capable of accurately predicting citation intention with respect to the three aforementioned labels for the KIM dataset is also of direct practical interest to the KIM project and the materials science community, as a whole.

\section{Additional Experimental Results} 

\subsection{Effectiveness of the TRL Method} \label{sec:add_exp_res}

\begin{figure*}[h!]
\centering
\includegraphics[width=\textwidth]{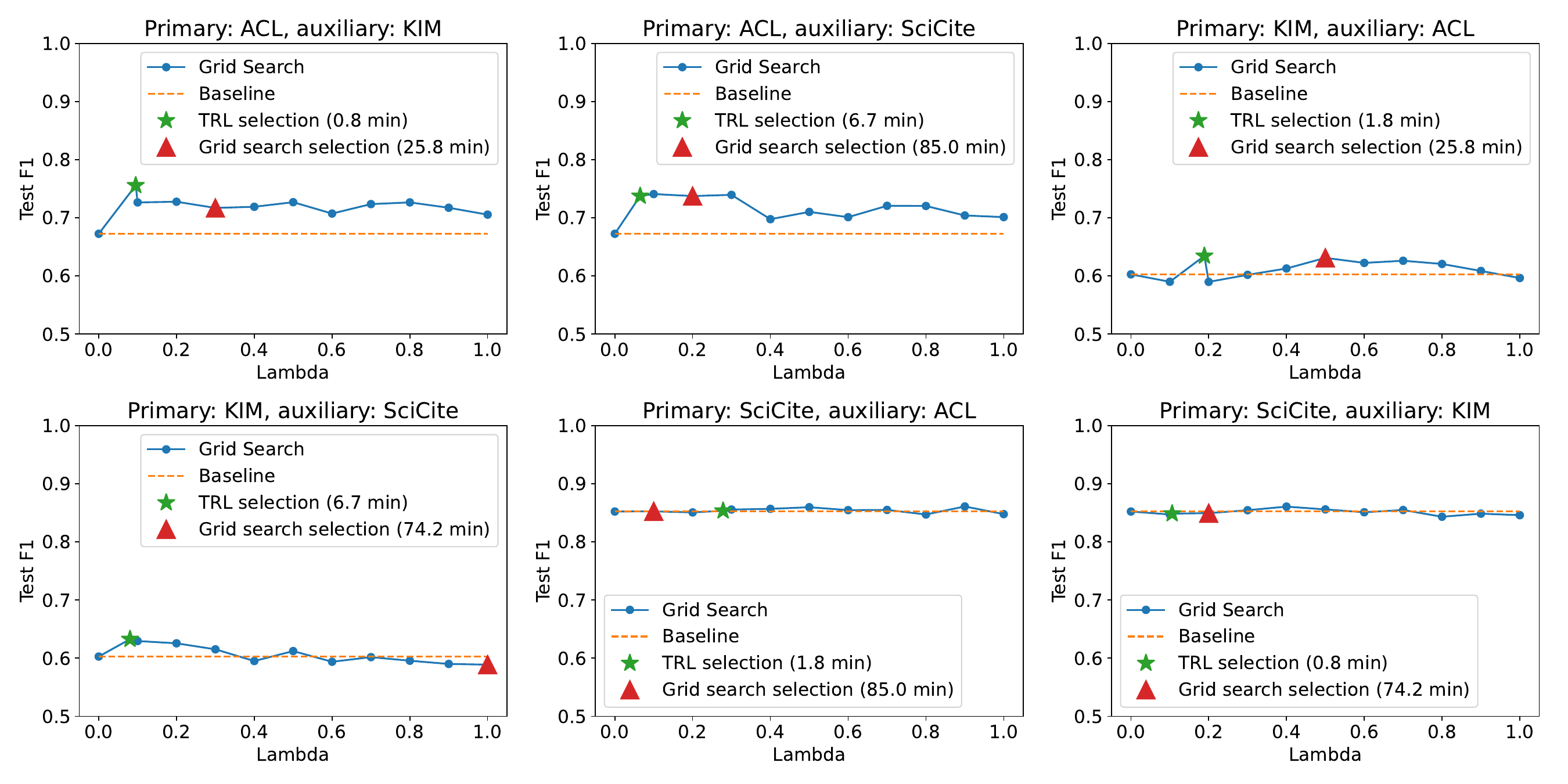}
\caption{SciBERT performance of all binary combinations of primary and auxiliary datasets with different value of $\lambda$s. The yellow line denotes the baseline performance of fine-tuning on only the primary dataset. The blue line denotes the performance of fine-tuning the primary and the auxiliary dataset with different $\lambda$s. The star and the triangle indicate the $lambda$ found by our TRL method and the grid search method, respectively. Time in the brackets indicates the GPU time needed for the method.}
\label{fig:trl_lambdas_bert}
\end{figure*}

In Figure~\ref{fig:trl_lambdas_bert}, we show SciBERT's performance of all binary combinations of primary and auxiliary datasets with different value of $\lambda$s. We have similar observations to the results for BERT. The fine-tuning performance is sensitive to the choice of $\lambda$s. 
Our TRL method can effectively identify $\lambda$s that improve the performance of the primary datasets and performs on par or better than those selected by the grid search method. 

\subsection{Multi-Dataset Fine-tuning}

We explore all combinations of primary and auxiliary datasets and show the results in Table~\ref{tab:data_predictions}. 
In most of the cases, jointly fine-tuning PLMs on primary datasets with auxiliary datasets using our proposed MTL framework improves the PLMs' performance on the primary datasets.
In cases when no improvement is achieved, jointly fine-tuning performs on par with fine-tuning the primary dataset by itself.
In the table, we also observe a few cases (e.g., Primary: KIM, Auxiliary: SciCite, PLM: SciBERT) where adding auxiliary datasets degrades the PLMs' performance on the primary dataset. 
While auxiliary datasets can bring additional knowledge to PLMs, they may also cause distribution shift that degrades the performance of the PLMs.

\begin{table}
  \resizebox{0.5\textwidth}{!}{%
  \centering
  \begin{tabular}{llrrrr}
    \toprule
    Primary & Auxiliary & \multicolumn{2}{c}{BERT} & \multicolumn{2}{c}{SciBERT} \\
    & & Search & TRL & Search & TRL \\
    \midrule
    \multirow{4}{*}{ACL} & - & 57.44 & 57.44 & 67.25 & 67.25 \\
     & KIM & 62.16$\uparrow$ & 64.39$\uparrow$ & 71.70$\uparrow$ & 75.57$\uparrow$ \\
     & SciCite & 65.98$\uparrow$ & 66.32$\uparrow$ & 73.74$\uparrow$ & 73.75$\uparrow$ \\
     & KIM + SciCite & 63.91$\uparrow$ & 62.53$\uparrow$ & 71.78$\uparrow$ & 72.87$\uparrow$ \\
    \midrule
    \multirow{4}{*}{KIM} & - & 57.30 & 57.30 & 60.27 & 60.27 \\
     & ACL                 & 60.70$\uparrow$ & 62.00$\uparrow$ & 63.11$\uparrow$ & 63.44$\uparrow$ \\
     & SciCite             & 63.21$\uparrow$ & 62.74$\uparrow$ & 58.88$\downarrow$ & 63.26$\uparrow$ \\
     & ACL + SciCite       & 64.18$\uparrow$ & 62.98$\uparrow$ & 61.86$\uparrow$ & 64.56$\uparrow$ \\
    \midrule
    \multirow{4}{*}{SciCite} & - & 83.46 & 83.46 & 85.22 & 85.22 \\
     & ACL & 83.96$\uparrow$ & 83.98$\uparrow$ & 85.25$\uparrow$ & 85.35$\uparrow$ \\
     & KIM & 84.08$\uparrow$ & 83.48$\uparrow$ & 84.94$\downarrow$ & 84.86$\downarrow$ \\
     & ACL + KIM & 84.34$\uparrow$ & 84.00$\uparrow$ & 85.43$\uparrow$ & 84.55$\downarrow$ \\
  \bottomrule
\end{tabular}}
 \caption{Performance (Macro-F1) of the MTL fine-tuning approach with different combinations of primary and auxiliary datasets. Search indicates grid search while TRL indicates our proposed task relation learning method. We group the results by the primary dataset. Baseline results of each primary dataset are shown in the first row of each group.}
 \label{tab:data_predictions}
\end{table}

\section{In-Context Learning Prompts} \label{sec:icl}

We present detailed examples of the zero-shot and few-shot prompts that we use for in-context learning on GPT4. We access to GPT4, specifically gpt-4-0125-preview, through the OpenAI API. 



\subsection{Zero-shot Prompt} 

{\fontfamily{qcr}\selectfont
I want you to act as a research assistant with expertise in atomistic modeling. I will provide you with a piece of text from a scientific paper that cites another paper. You will classify the text into one of the following labels that indicate the intention of the citation: [Background, Method, Result]. The labels are defined as

"Background": "The citation states, mentions, or points to the background information giving more context about a problem, concept, approach, topic, or importance of the problem that is discussed in the present paper."

"Method": "The present paper uses a method, tool, approach or dataset that is proposed in the paper associated with the citation."

"Result": "The present paper compares its results/findings with the results/findings of the paper associated with the citation."

You will only respond with the predicted label. Below is the input text:

We used an active contour algorithm <CITED HERE> to segment the organs from 340 coronal slices over the two patients.}

\subsection{Few-shot Prompt} 

{\fontfamily{qcr}\selectfont
I want you to act as a research assistant with expertise in atomistic modeling. I will provide you with a piece of text from a scientific paper that cites another paper. You will classify the text into one of the following labels that indicate the intention of the citation: [Background, Method, Result]. The labels are defined as

"Background": "The citation states, mentions, or points to the background information giving more context about a problem, concept, approach, topic, or importance of the problem that is discussed in the present paper."

"Method": "The present paper uses a method, tool, approach or dataset that is proposed in the paper associated with the citation."

"Result": "The present paper compares its results/findings with the results/findings of the paper associated with the citation."

Here are some examples:

Example: We used an active contour algorithm <CITED HERE> to segment the organs from 340 coronal slices over the two patients.

Output: Method

Example: The remnant of the total plasma membranes after extraction of caveolae is called bulk plasma membranes <CITED HERE> (Fig.

Output: Background

Example: More examples of contradictory results have been observed in bovines; some reports <CITED HERE> indicated a significant decrease in blastocyst

Output: Result

You will only respond with the predicted label. Below is the input text:

Following <CITED HERE> and Koo and Collins (2010), before training we transform the training set trees to be the best achievable within the model class (i.e., the closest projective tree or 1-Endpoint-Crossing tree).}

\section{Experimental Settings} \label{sec:exp_setting}

We use the Adam optimizer~\cite{kingma2014adam} to minimize the cross-entropy loss in all our pre-training and fine-tuning tasks. The batch size is set to 32. For fine-tuning tasks, we set the learning rate to be 5e-5 and use a slated triangular scheduler~\cite{howard2018slanted} to first warm up and then decrease the learning rate linearly. The model is fine-tuned for 10 epochs and evaluated on the validation set after every epoch. 
For the ACL and the SciCite datasets, we use the original train-test split and use 15\% of the training set as validation set. For the KIM dataset, we randomly split the dataset into train, validation, and test sets with a 70\%/15\%/15\% ratio. 
Test performance is reported on the checkpoint that performs the best on the validation set. We report the macro-F1 score as the evaluation metric.
The macro-F1 scores that we report in this paper are averaged numbers of five independent runs. All experiments are conducted on a machine with an Intel(R) Core(TM) i9-10900F CPU and an Nvidia RTX 3090 GPU.
Our methods and experiments are implemented using PyTorch~\cite{paszke2019pytorch}. For experiments including BERT and SciBERT, we use the implementation and pre-trained weights from the transformers library.\footnote{\url{https://github.com/huggingface/transformers}} 

\begin{figure}[t]
\centering
\includegraphics[width=0.5\textwidth]{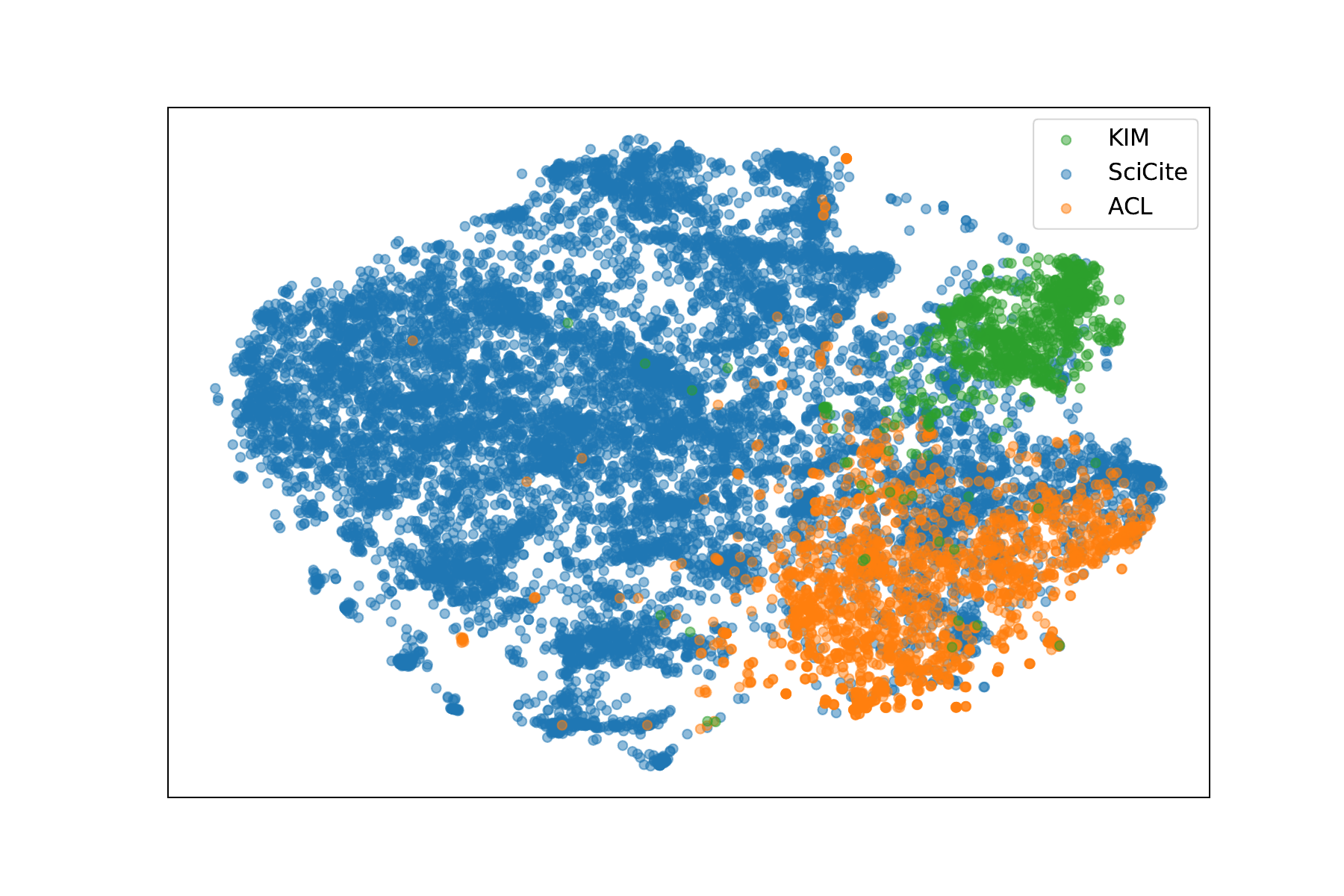}
\caption{T-SNE visualization of the citation contexts in different datasets. Citation contexts are encoded by SciBERT using the CLS readout function.}
\label{fig:tsne}
\end{figure}

\section{Visualization of Citation Contexts}

In Figure~\ref{fig:tsne}, we show the t-SNE~\cite{van2008tsne} visualization of the citation contexts in the three datasets. We use SciBERT and the CLS readout function to convert the contexts to latent embeddings. We observe that the citation contexts of different datasets form into clusters because the fields of the papers are different, but there are significant overlaps between the clusters. It is reasonable to assume that the datasets share an input space.